\documentclass[sigconf]{aamas}

\setcopyright{none}
\renewcommand\footnotetextcopyrightpermission[1]{}

\usepackage{amsmath}
\usepackage{amsxtra}
\usepackage{amstext}
\usepackage{amsfonts}
\usepackage{amssymb}
\usepackage{amsthm}

\usepackage{booktabs}
\usepackage{makecell}
\usepackage{siunitx}

\usepackage{graphicx}
\usepackage{subcaption}

\usepackage{algorithm}
\usepackage[noend]{algpseudocode}

\usepackage{tikz}

\usepackage{todonotes}
\usepackage{lipsum}
\usepackage{xcolor}  
\usepackage{siunitx}

\newcommand{\vect}[1]{\boldsymbol #1}
\newcommand{\mat}[1]{\boldsymbol{\mathrm #1}}

\newcommand{\A}{\mathcal{A}}
\newcommand{\D}{\mathcal{D}}
\newcommand{\F}{\mathcal{F}}
\newcommand{\I}{\mathcal{I}}
\newcommand{\M}{\mathcal{M}}
\newcommand{\X}{\mathcal{X}}

\newcommand{\Z}{\mathcal{Z}}

\newcommand{\EE}[2][]{\mathbb{E}_{#1}\left[#2\right]} 

\DeclareMathOperator{\KL}{\mathbb{K}\mathbb{L}}

\begin{document}

\title[An approach for cross-modality transfer in reinforcement
learning]{Playing Games in the Dark: An approach for cross-modality
  transfer in reinforcement learning} 

\author{
  {Rui Silva}\textsuperscript{\rm 1,2,3},
  {Miguel Vasco}\textsuperscript{\rm 1,2},
  {Francisco S. Melo}\textsuperscript{\rm 1,2},
  {Ana Paiva}\textsuperscript{\rm 1,2},
  {Manuela Veloso}\textsuperscript{\rm 3}\\
  \textsuperscript{1}INESC-ID, Lisboa, Portugal
  \textsuperscript{2}Instituto Superior T\'{e}cnico, University of Lisbon, Portugal\\
  \textsuperscript{3}Carnegie Mellon University, Pittsburgh, USA\\
}

\renewcommand{\shortauthors}{R. Silva et al.}

\begin{abstract}
  In this work we explore the use of latent representations obtained
  from multiple input sensory modalities (such as images or sounds) in
  allowing an agent to learn and exploit policies over different
  subsets of input modalities. We propose a three-stage architecture
  that allows a reinforcement learning agent trained over a given
  sensory modality, to execute its task on a different sensory
  modality---for example, learning a visual policy over image inputs,
  and then execute such policy when only sound inputs are
  available. We show that the generalized policies achieve better
  out-of-the-box performance when compared to different
  baselines. Moreover, we show this holds in different OpenAI gym and
  video game environments, even when using different multimodal
  generative models and reinforcement learning algorithms.
\end{abstract}


\maketitle
\section{Introduction}
\label{sec:introduction}

Recent works have shown how low-dimensional representations captured
by generative models can be successfully exploited in reinforcement
learning (RL) settings. Among others, these generative models have
been used to learn low-dimensional latent representations of the state
space to improve the learning efficiency of RL
algorithms~\cite{zhang2018arxiv,gelada2019arxiv}, or to allow the
generalization of policies learned on a source domain to other target
domains~\cite{finn2016icra,higgins2017icml}. The DisentAngled
Representation Learning Agent (DARLA) approach~\cite{higgins2017icml},
in particular, builds such latent representations using variational
autoencoder (VAE) methods~\cite{kingma2013arxiv,rezende2014arxiv}, and
shows how learning disentangled features of the observed environment
can allow an RL agent to learn a policy robust to some shifts in the
original domain.

\begin{figure}[t]
  \centering
  \includegraphics[width=8.5cm]{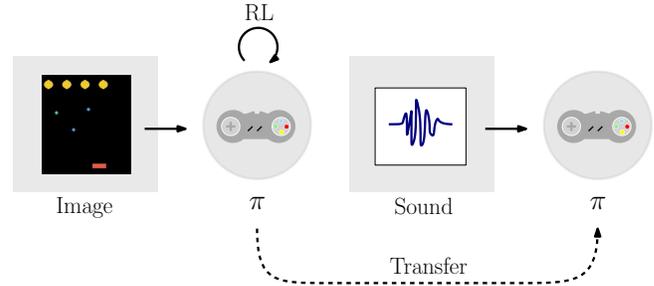}
  \caption{Concrete scenario where a policy trained over
    one input modality (game videoframes) is transferred to a
    different modality (game sound).}
  \label{fig:example}
\end{figure}

In this work, we explore the application of these latent
representations in capturing different input sensory modalities to be
considered in the context of RL tasks. We build upon recent work that
extends VAE methods to learn joint distributions of multiple
modalities, by forcing the individual latent representations of each
modality to be similar~\cite{suzuki2016arxiv,yin2017aaai}. These
multimodal VAEs allow for \emph{cross-modality inference}, replicating
more closely what seems to be the nature of the multimodal
representation learning performed by
humans~\cite{damasio1989cognition,meyer2009tns}. Inspired by these
advances, we explore the impact of such multimodal latent
representations in allowing a reinforcement learning agent to learn
and exploit policies over different input modalities. Among others, we
envision, for example, scenarios where reinforcement learning agents
are provided the ability of learning a visual policy (a policy learned
over image inputs), and then (re-)using such policy at test time when
only sound inputs are available. Figure~\ref{fig:example} instantiates
such example to the case of video games---a policy is learned over
images and then re-used when only the game sounds are available,
\emph{i.e.,} when playing ``in the dark''.

To achieve this, we contribute an approach for multimodal transfer
reinforcement learning, which effectively allows an RL agent to learn
robust policies over input modalities, achieving better out-of-the-box
performance when compared to different baselines. We start by first
learning a generalized latent space over the different input
modalities that the agent has access to. This latent space is
constructed using a multimodal generative model, allowing
the agent to establish mappings between the different modalities---for
example, \emph{``which sounds do I typically associate with this
  visual sensory information''}. Then, in the second step, the RL
agent learns a policy directly on top of this latent
space. Importantly, during this training step, the agent may only have
access to a subset of the input modalities (say, images but not
sound). In practice, this translates in the RL agent learning a policy
over a latent space constructed relying only on some
modalities. Finally, the transfer occurs in the third step, where, at
test time, the agent may have access to a different subset of
modalities, but still perform the task using the same policy. These
results hold consistently across different OpenAI
Gym~\cite{brockman2016arxiv} and Atari-like~\cite{bellemare2013jair}
environments. This is the case even when using different multimodal
generative models~\cite{yin2017aaai} and reinforcement
learning algorithms~\cite{mnih2015nature,lillicrap2015arxiv}.

\begin{figure*}[t]
  \centering
  \begin{subfigure}[t]{0.16\linewidth}
    \centering
    \includegraphics[height=4.3cm]{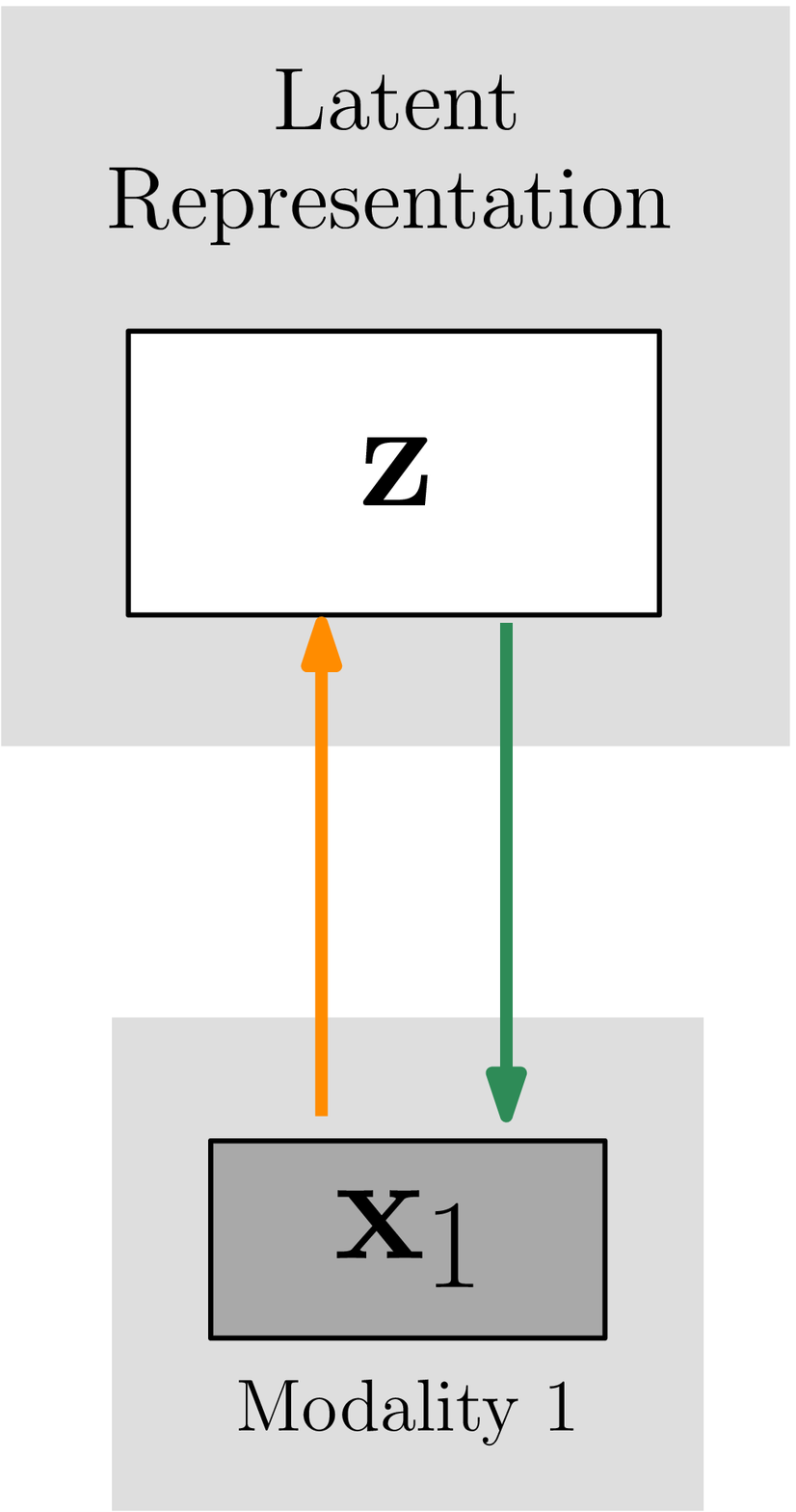}
    \caption{}
    \label{fig:vae_model}
  \end{subfigure}
  %
  %
  \begin{subfigure}[t]{0.31\linewidth}
    \centering
    \includegraphics[height=4.3cm]{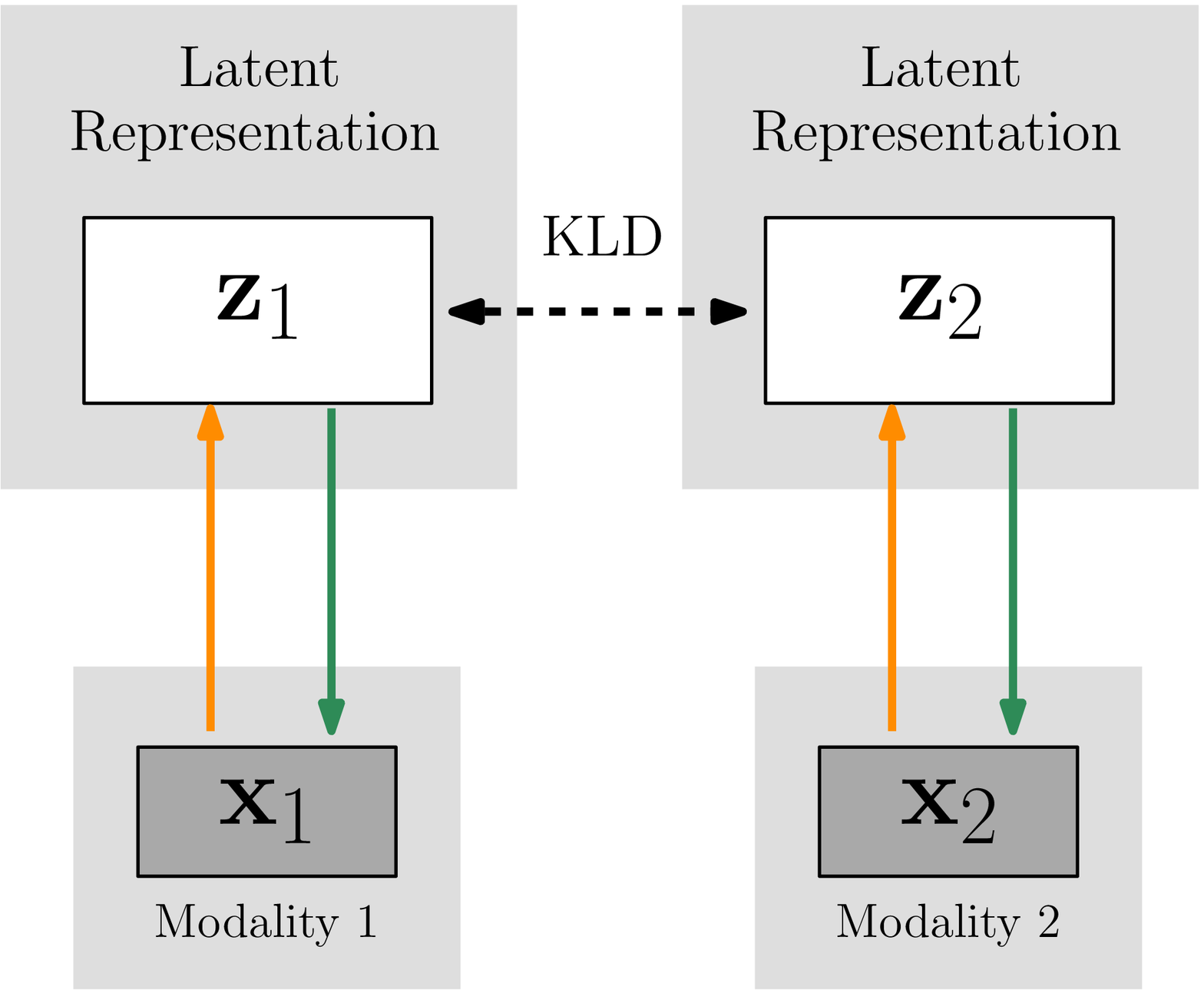}
    \caption{}
    \label{fig:avae_model}
  \end{subfigure}
  %
  %
  \begin{subfigure}[t]{0.52\linewidth}
    \centering
    \includegraphics[height=4.3cm]{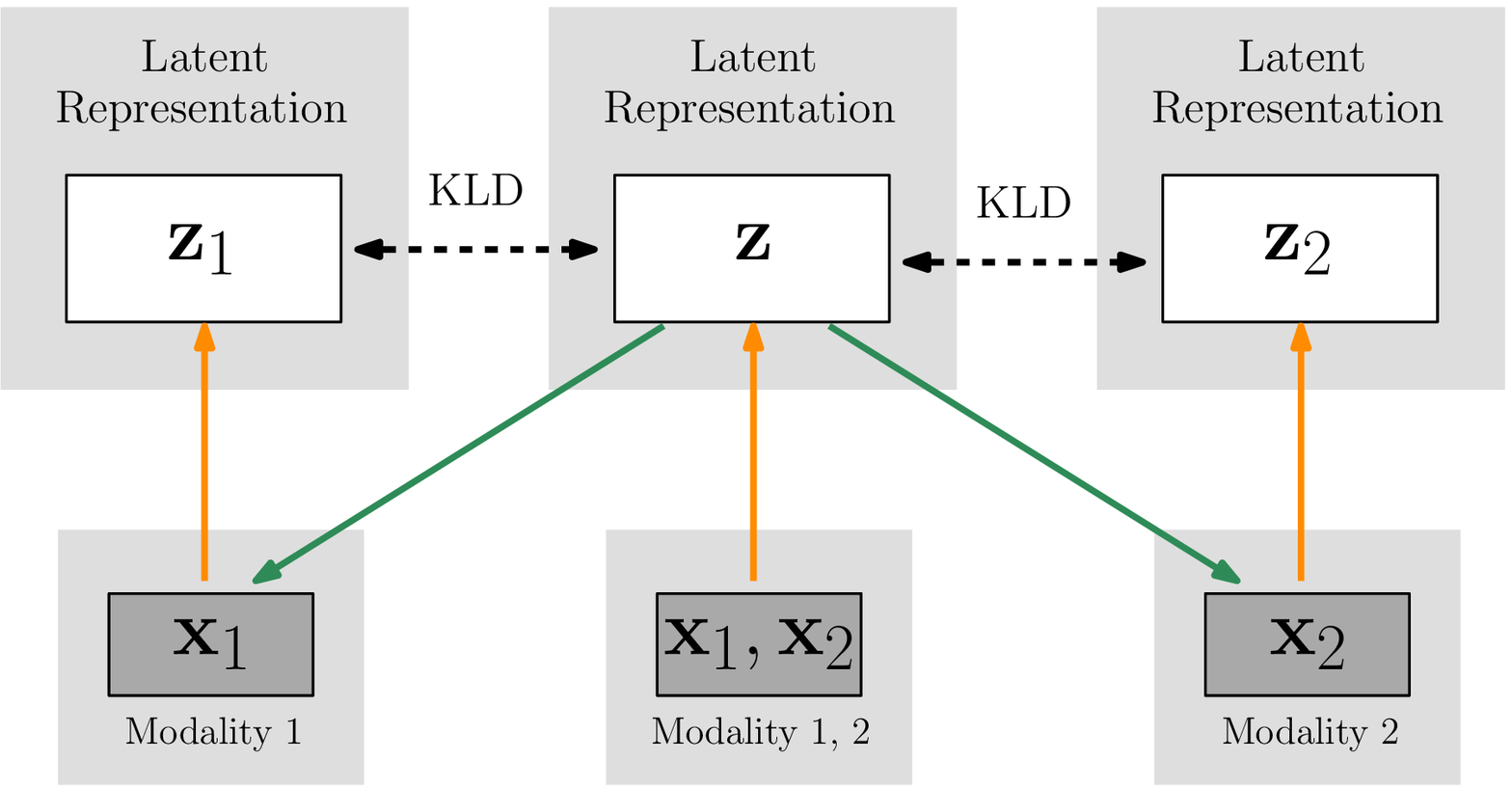}
    \caption{}
    \label{fig:jmvae_model}
  \end{subfigure}
  \caption{Networks of different generative models, highlighting the
    models' data encoding (orange) and decoding (green) pipelines. The
    similarity constraints imposed by their training procedures are
    presented in dashed lines. \ref{fig:vae_model}) The VAE model
    learns a latent representation of the data distribution of a
    single modality. \ref{fig:avae_model}) On the other hand, the AVAE
    model extends the previous framework to account for multiple
    modalities, allowing for cross-modality
    inference. \ref{fig:jmvae_model}) Finally, the JMVAE model learns
    a representation of both modalities, allowing for both single and
    joint modality reconstruction, and cross-modality inference.}
  \label{fig:deep_generative_models}
\end{figure*}

The third and last step unveils what sets our work apart from the
existing literature. By using (single-modality) VAE methods, the
current state-of-art approaches implicitly assume that the source and
target domains are characterized by similar inputs, such as raw
observations of a camera. In these approaches, the latent space is
used to capture isolated properties (such as colors or shapes) that
may vary throughout the tasks. This is in contrast with our approach,
where the latent space is seen as a mechanism to create a mapping
between different input modalities.

The remainder of the paper is structured as follows. We start in
Section~\ref{sec:preliminaries} by introducing relevant background and
related work on generative models and reinforcement learning. Then, in
Section~\ref{sec:approach} we introduce our approach to multimodal
transfer reinforcement learning, and evaluate it in
Section~\ref{sec:experimental evaluation}. We finish with some final
considerations in Section~\ref{sec:conclusions}.

\section{Preliminaries}
\label{sec:preliminaries}

This section introduces required background on deep generative models
and deep reinforcement learning.

\subsection{Deep Generative Models}

\subsubsection{Variational Autoencoders}

Deep generative models have shown great promise in learning
generalized representations of data. For single-modality data, the
variational autoencoder model (VAE) is widely used. The VAE
model~\cite{kingma2013arxiv} learns a joint distribution
$p_{\theta}(\vect{x}, \vect{z})$ of data $\vect{x}$, which is
generated by a latent variable $\vect{z}$.  Figure~\ref{fig:vae_model}
depicts this model. The latent variable is often of lower
dimensionality in comparison with the modality itself, and acts as the
representation vector in which data is encoded.

The joint distribution takes the form
$p_{\theta}(\vect{x}, \vect{z}) = p_{\theta}(\vect{x}\, \vert\,
\vect{z}) \, p(\vect{z})$, where $p(\vect{z})$ (the
\textit{prior} distribution) is often a unitary Gaussian
($\vect{z} \sim \mathcal{N}(\vect{0}, \mat{I})$). The
generative distribution $p_{\theta}(\vect{x}\, \vert\, \vect{z})$,
parameterized by $\theta$, is usually composed with a simple
likelihood term (e.g. Bernoulli or Gaussian).

The training procedure of the VAE model involves the maximization of
the evidence likelihood $p(\vect{x})$, by marginalizing over the
latent variable and resorting to an inference network
$q_{\phi}(\vect{z}\vert\vect{x})$ to approximate the posterior
distribution. We obtain a lower-bound on the log-likelihood of the
evidence (ELBO)
$\log p(\vect{x}) \geq \mathcal{L}_{\text{VAE}}(\vect{x})$, with
\begin{equation*}
  \mathcal{L}_{\text{VAE}}(\vect{x})
  =
  \lambda\, \EE[q_{\phi}(\vect{z}\vert\vect{x})]
  {\log p_{\theta} (\vect{x}\vert\vect{z})}  - \, \beta\, \KL \left[q_{\phi}(\vect{z} \vert \vect{x}) \, \| \, p(\vect{z})\right],
\end{equation*}
where the Kullback-Leibler divergence term
$\KL \left[q_{\phi}(\vect{z} \vert \vect{x}) \,\|\,
  p(\vect{z})\right]$ promotes a balance between the latent channel's
capacity and the encoding process of data. Moreover, in the model's
training procedure, the hyperparameters $\lambda$ and $\beta$ weight
the importance of reconstruction quality and latent space
independence, respectively. The optimization of the ELBO is performed
resorting to gradient-based methods.

\subsubsection{Multimodal Variational Autoencoders}

VAE models have been extended in order to perform inference across
different modalities. The Associative Variational Autoencoder (AVAE)
model~\cite{yin2017aaai}, depicted in Figure~\ref{fig:avae_model}, is
able to learn a common latent representation of two modalities
($\vect{x}, \vect{y}$). It does so by imposing a similarity
restriction on the separate single-modality latent representations
($\vect{z}_{x}, \vect{z}_{y}$), employing a KL divergence term on the
ELBO of the model:
\begin{equation*}
  \mathcal{L}_{\text{AVAE}}(\vect{x}, \vect{y})
  =
  \mathcal{L}_{\text{VAE}}(\vect{x}) + \mathcal{L}_{\text{VAE}}(\vect{y}) - \,\alpha \KL^{\star} \left[q_{\phi}(\vect{z}_{x} \vert \vect{x}) \, \| \, q_{\phi}(\vect{z}_{y} \vert \vect{y})\right]
\end{equation*}
where $\KL^{\star} \left[p \, \| \,q \right]$ is the symmetrical
Kullback-Leibler between two distributions $p$ and $q$, and $\alpha$
is a constant that weights the importance of keeping similar latent
spaces in the training procedure~\cite{yin2017aaai}. We note that each
modality is associated with a different encoder-decoder
pair. Moreover, the encoder and the decoder can be implemented as
neural networks with different architectures.

Other models aim at learning a joint distribution of both modalities
$p_\theta(\vect{x}, \vect{y})$. Examples include the Joint Multimodal
Variational Autoencoder (JMVAE)~\cite{suzuki2016arxiv} or the
Multi-Modal Variational Autoencoder
(M$^2$VAE)~\cite{korthals2019arxiv}. These generative models are able
to build a representation space of both modalities simultaneously
while maintaining similarity restrictions with the single-modality
representations, as shown in the JMVAE model presented in
Figure~\ref{fig:jmvae_model}.

However, a fundamental feature of all multimodal generative models is
the ability to perform \emph{cross-modality inference}, that is the
ability to input modality-specific data, encode the corresponding
latent representation, and, from that representation, generate data
of a different modality. This is possible due to the forced
approximation of the latent representations of each modality, and the
process follows the orange and green arrows in
Figure~\ref{fig:deep_generative_models}.

\subsection{Reinforcement Learning}

Reinforcement learning (RL) is a framework for optimizing the
behaviour of an agent operating in a given environment. This framework
is formalized as a Markov decision process (MDP)---a tuple
$\M = (\X,\A,P,r,\gamma)$ that describes a sequential decision problem
under uncertainty. $\X$ and $\A$ are the state and action spaces,
respectively, and both are known by the agent. When the agent takes an
action $a\in\A$ while in state $x\in\X$, the world transitions to
state $y\in\X$ with probability $P(y\mid x,a)$ and the agent receives
an immediate reward $r(x,a)$. Typically, functions $P$ and $r$ are
unknown to the agent.  Finally, the discount factor
$\gamma \in [0, 1)$ sets the relative importance of present and future
rewards.

Solving the MDP consists in finding an optimal policy $\pi^*$---a
mapping from states to actions---which ensures that the agent collects
as much reward as possible. Such policy can be found from the optimal
$Q$-function, which is defined recursively for every state action pair
$\left( x, a \right) \in \X \times \A$ as
\begin{equation*}
  Q^*(x,a)
  =
  r(x, a)
  +
  \gamma
  \sum_{y\in\X}
  P(y \mid x, a)
  \max_{a'\in\A} Q^*(y, a').
\end{equation*}
Multiple methods can be used in computing this
function~\cite{sutton1998reinforcement}, for example
$Q$-learning~\cite{watkins89phd}.

More recently, research has geared towards applying deep learning
methods in RL problems, leading to new methods. For example, Deep $Q$
Network (DQN) is a variant of the $Q$-learning algorithm that uses a
deep neural network to parameterize an approximation of the
$Q$-functions $Q(x, a; \theta)$, with parameters $\theta$. DQN assumes
discrete action spaces $\A$, and has been proved suitable for learning
policies that beat Atari games~\cite{mnih2015nature}. Continuous
action spaces require specialized algorithms. For example, Deep
Deterministic Policy Gradient (DDPG) is an actor-critic, policy
gradient algorithm that can deal with continuous action spaces, and
has been shown to perform well in complex control
tasks~\cite{lillicrap2015arxiv}.

\section{Multimodal Transfer Reinforcement Learning}
\label{sec:approach}

\begin{figure}[t]
  \centering
  \includegraphics[width=4cm]{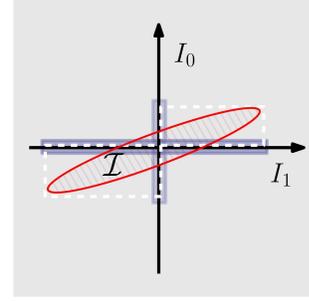}
  \caption{Connection between two abstract input modalities $I_1$ and
    $I_2$, and corresponding impact in the agent's perceptual
    information. The elliptic surface $\I$ depicts the complete
    perceptual space the agent can perceive with both modalities. The
    colored projections in the axes depict the (reduced) perceptions
    of the agent when using single modalities.}
  \label{fig:perceptual information}
\end{figure}

Consider an agent facing a sequential decision problem described as an
MDP $\M = \left( \X, \mathcal{A}, P, r, \gamma \right)$. This agent is
endowed with a set $\left\lbrace I_1, I_2, \dots, I_N \right\rbrace$
of $N$ different input modalities, which can be used in perceiving the
world and building a possibly partial observation of the current state
$x \in \X$. Different modalities may provide more, or less, perceptual
information than others. Some modalities may be redundant
(\emph{i.e.}, provide the same perceptual information) or complement
each other (\emph{i.e.}, jointly provide more
information). Figure~\ref{fig:perceptual information} provides an
abstract illustration of the connection between different input
modalities, and corresponding impact in the state space that can be
perceived by the agent.

Our goal is for the agent to learn a policy while observing only a
subset of input modalities $\boldsymbol{I}_\mathrm{train}$, and then use that same
policy when observing a possibly different subset of modalities,
$\boldsymbol{I}_\mathrm{test}$, with as minimal performance degradation as
possible.

Our approach consists of a three-stages pipeline:
\begin{enumerate}
\item
  \emph{Learn a perceptual model of the world.}
\item
  \emph{Learn to act in the world.}
\item
  \emph{Transfer policy.}
\end{enumerate}

We now discuss each step in further detail.

\subsection{Learn a perceptual model of the world}
\label{subsec:generative model}

Let $\I$ denote the Cartesian product of input modalities,
$\I = I_1 \times I_2 \times \dots \times I_N$. Intuitively, we can
think of $\I$ as the complete perceptual space of the agent. We write
$\vect{i}$ to denote an element of $\I$. Figure~\ref{fig:map
  modalities} depicts an example on a game, where the agent can have
access to two modalities, $I_\mathrm{image}$ and $I_\mathrm{sound}$,
corresponding to visual and sound information.

\begin{figure*}[t]
  \centering
  \begin{subfigure}[t]{0.49\textwidth}
    \centering
    \includegraphics[width=8.20cm]{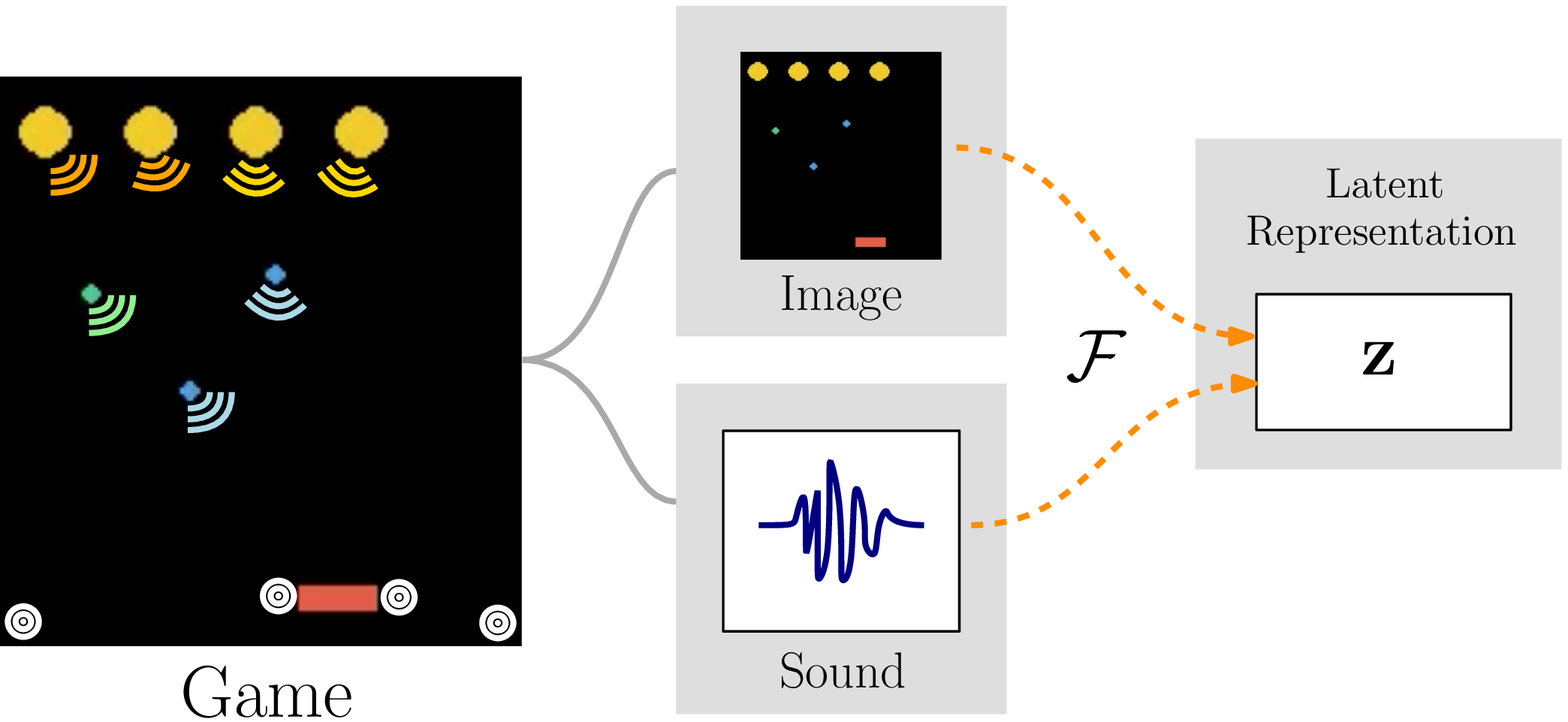}
    \caption{}
    \label{fig:map modalities}
  \end{subfigure}
  \hfill
  \begin{subfigure}[t]{0.49\textwidth}
    \centering
    \includegraphics[width=8.10cm]{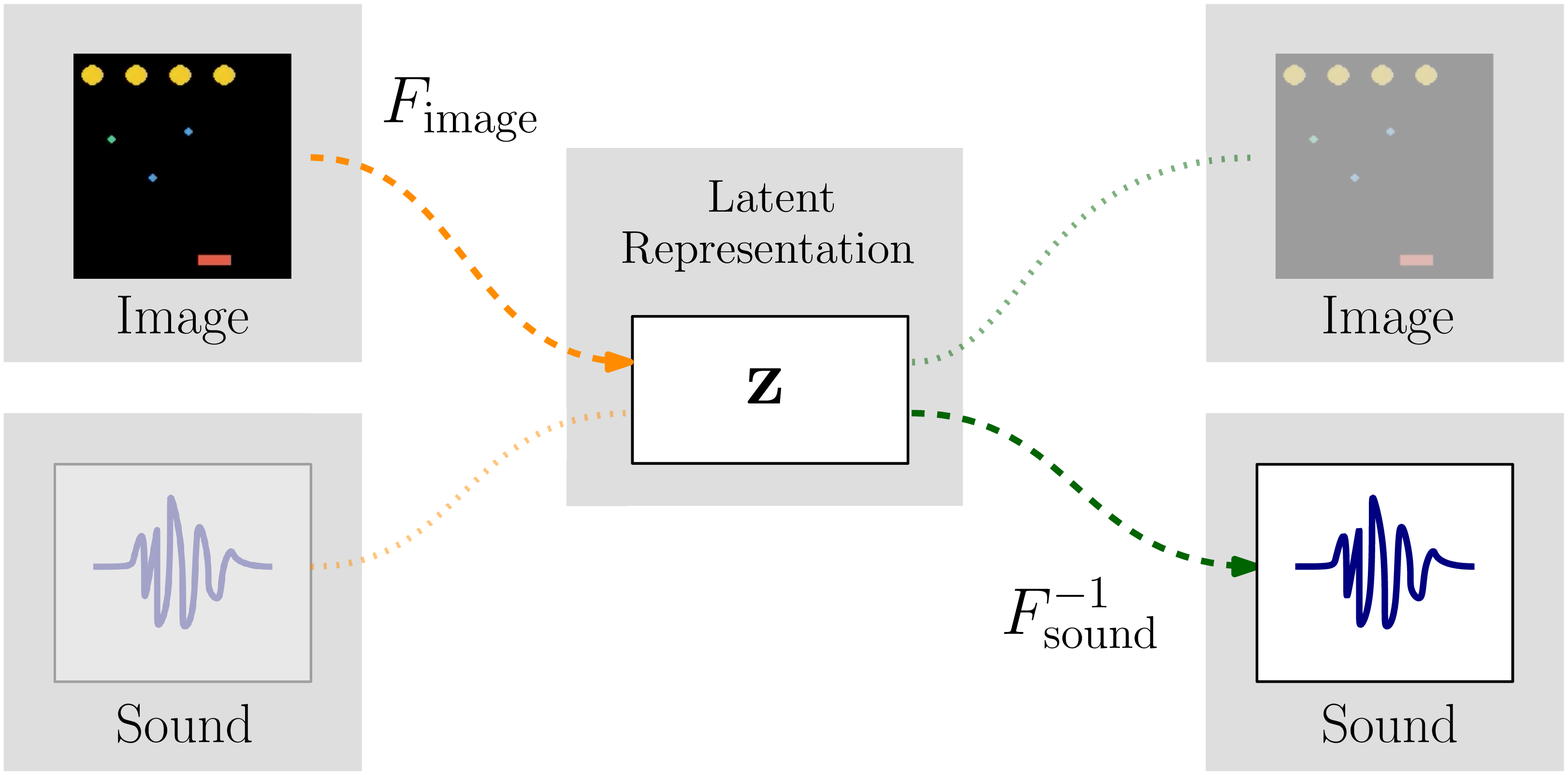}
    \caption{}
    \label{fig:cross modality}
  \end{subfigure}
  \caption{\ref{fig:map modalities}) Each time step of a game includes
    visual and sound components that are intrinsically coupled. This
    coupling  can be encoded in a latent representation using the
    family of latent maps $\F$.
    \ref{fig:cross modality}) shows how the multimodal latent
    representation can be used in inferring the sound associated with
    a given image, using the image latent map $F_\mathrm{image}$ and
    sound inverse latent map $F_{\mathrm{sound}}^{-1}$.}
  \label{fig:map and cross modality}
\end{figure*}

At each moment $t$, the agent may not have access to the complete
perception $\vect{i}(t)\in\I$, but only to a partial view
thereof. Following our discussion in Section~\ref{sec:introduction},
we are interested in learning a multimodal latent representation of
the perceptions in $\I$. Such representation amounts to a set of
latent mappings $\F=\{F_1,\ldots,F_L\}$. Each map $F_\ell$ takes the
form $F_\ell:\mathrm{proj}_\ell\to\Z$, where $\Z$ is a common latent
space and $\mathrm{proj}_\ell$ projects $\I$ to some subspace of $K$
modalities,
$\I_\ell=I_{\ell_1}\times I_{\ell_2}\times\ldots\times I_{\ell_K}$. In
Figure~\ref{fig:map modalities} the set of mappings $\F$ is used to
compute a latent representation $\vect{z}$ from sound and image data.

To learn such mappings, we start by collecting a dataset
of $M$ examples of simultaneous sensorial information:
\begin{equation*}
  \D(\I)
  =
  \big\{
    \vect{i}^{(1)},\ldots,\vect{i}^{(M)}
  \big\}.
\end{equation*}
We then follow an unsupervised learning approach, and train a
multimodal VAE on dataset $\D(\I)$ to learn a generalized latent space
over the agent's input modalities. The latent mappings in $\F$
correspond to the encoders of the VAE model, while the decoders can be
seen as a set of inverse latent mappings,
$\F^{-1}=\{F^{-1}_1,\ldots,F^{-1}_L\}$ that allow for modality
reconstruction and cross-modality inference. Figure~\ref{fig:cross
  modality} depicts an example of how the multimodal latent space can
be used for performing cross-modality inference of sound data given an
image input using the modality-specific maps.

The collection of the initial data needed to generate $\D(\I)$ can be
easier/harder depending on the complexity of the task. In
Section~\ref{sec:experimental evaluation} we discuss mechanisms to
perform this.

\subsection{Learn to act in the world}
\label{subsec:reinforcement learning}

After learning a perceptual model of the world, the agent then learns
how to perform the task. We follow a reinforcement learning approach
to learn an optimal policy for the task described by MDP $\M$. During
this learning phase, we assume the agent may only have access to a
subset of input modalities $\boldsymbol{I}_\mathrm{train}$.
As a result, during its interaction with the environment, the agent
collects a sequence of triplets
\begin{equation*}
  \left\lbrace
    \left(
      \vect{i}_\mathrm{train}^{(0)},
      a^{(0)},
      r^{(0)}
    \right),
    \left(
      \vect{i}_\mathrm{train}^{(1)},
      a^{(1)},
      r^{(1)}
    \right),
    \dots
  \right\rbrace,
\end{equation*}
where $\vect{i}_\mathrm{train}^{(t)}$, $a^{(t)}$, $r^{(t)}$ correspond
to the perceptual observations, action executed, and rewards obtained
at timestep $t$, respectively.

However, our reinforcement learning agent does not use this sequence
of triplets directly. Instead, it pre-processes the perceptual
observations using the previously learned latent maps
$\F$ to encode the multimodal latent state at each
time step as
$
\vect{z}^{(t)}
=
F_\mathrm{train}
\left(
  \vect{i}_\mathrm{train}^{(t)}
\right)
$,
where $F_\mathrm{train}\in\F$ maps $I_\mathrm{train}$ into $\Z$. In practice, the RL agent uses a sequence of triplets
\begin{equation*}
  \left\lbrace
    \left(
      \vect{z}^{(0)},
      a^{(0)},
      r^{(0)}
    \right),
    \left(
      \vect{z}^{(1)},
      a^{(1)},
      r^{(1)}
    \right),
    \dots
  \right\rbrace
\end{equation*}
to learn a policy $\pi : \Z \to \A$, that maps the latent states to
actions. Any continuous-state space reinforcement learning algorithm
can be used to learn this policy $\pi$ over the latent states. These
latent states are encoded using the generative model trained in the
previous section, and as such, the weights of this model should remain
frozen during the RL training.

\subsection{Transfer policy}
\label{subsec:transfer}

The transfer of policies happens once the agent has learned how to
perceive and act in the world. At this time, we assume the agent may
now have access to a subset of input modalities
$\boldsymbol{I}_\mathrm{test}$, potentially different from
$\boldsymbol{I}_\mathrm{train}$, \emph{i.e.}, the set of modalities it
used in learning the task policy $\pi$. As a result, during its
interaction with the environment, at each time step $t$, the agent will now
observe perceptual information $\vect{i}_\mathrm{test}^{(t)}$.

In order to reuse the policy $\pi$, the agent starts by pre-processing
this perceptual observation, again using the set of maps $\F$
previously trained, but now generating a latent state
$\vect{z}^{(t)} = F_\mathrm{test} \left( \vect{i}_\mathrm{test}^{(t)}
\right)$, where $F_\mathrm{test}\in\F$ now maps $I_\mathrm{test}$ into $\Z$. Since policy $\pi$ maps the latent space $\Z$ to the action
space $\A$, it can now be used directly to select the optimal action at the new
state $\vect{z}^{(t)}$.

\begin{figure}[b]
  \centering
  \includegraphics[width=8cm]{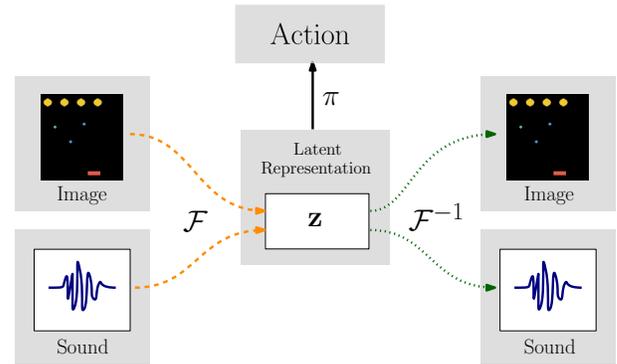}
  \caption{Summary of the three-steps approach for cross-modality
    transfer in reinforcement learning. The first step learns a
    perceptual model of the world, described by the latent mappings
    $\F$ (and corresponding inverses), which map perceptions to a
    common latent space $\Z$. In the second step, the agent learns a
    policy $\pi$ that maps the latent space to actions, with an RL
    approach using observations from a given subset of input
    modalities. The third step concerns the reuse off the same policy
    $\pi$, assuming new observations from a potentially different
    subset of modalities. This is possible by first encoding the new
    observations in the multimodal latent space.}
  \label{fig:learn_to_act}
\end{figure}

Effectively, the agent is reusing a policy $\pi$ that was learned over
a (possibly) different set of input modalities, with no additional
training. This corresponds to a zero-shot transfer of
policies. Figure~\ref{fig:learn_to_act} summarizes the three-steps
pipeline hereby described.

\section{Experimental Evaluation}
\label{sec:experimental evaluation}

We evaluate and analyze the performance of our approach on different
scenarios of increasing complexity, not only on the task but also on
the input modalities. We start by considering a modified version of
the \textsc{pendulum} environment from OpenAI gym, with a simple sound
source. Then, we consider \textsc{hyperhot}, a \textsc{space
  invaders}-like game that assesses the performance of our approach in
scenarios with more complex and realistic generation of sounds.

\begin{figure}[t]
  \centering
  \includegraphics[width=3cm]{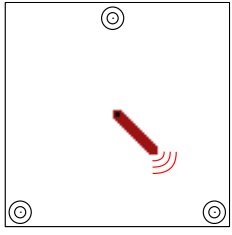}

  \caption{Visual and sound perceptual information in the
    \textsc{pendulum} scenario. The tip of the pendulum emits a
    frequency that is received by three microphones placed at the
    bottom left and right ($bl,br$) and middle top $(mt)$.}
  \label{fig:pendulum}
\end{figure}

\subsection{\textsc{pendulum}}

We consider a modified version of the \textsc{pendulum} environment
from OpenAI gym---a classic control problem, where the goal is to
swing the pendulum up so it stays upright. We modify this environment
so that the observations include both an image and a sound
component. For the sound component, we assume that the tip of the
pendulum emits a constant frequency $f_0$, which is received by a set
of $S$ sound receivers
$\left\lbrace \rho_1, \dots, \rho_S
\right\rbrace$. Figure~\ref{fig:pendulum} depicts this scenario, where
the pendulum and its sound are in red, and the sound receivers
correspond to the circles.

Formally, we let $\I = I_{\mathrm{image}} \times I_{\mathrm{sound}} $
denote the complete perceptual space of the agent. The visual input
modality of the agent, $I_{\mathrm{image}}$, consists of the raw image
observation of the environment. On the other hand, the sound input
modality, $I_{\mathrm{sound}}$, consists of the frequency and
amplitude received by each of the $S$ microphones of the
agent. Moreover, both image and sound observations may be stacked to
account for the dynamics of the scenario.

\begin{figure}[t]
  \centering
  \begin{subfigure}[t]{0.49\linewidth}
    \centering
    \includegraphics[width=3.57cm]{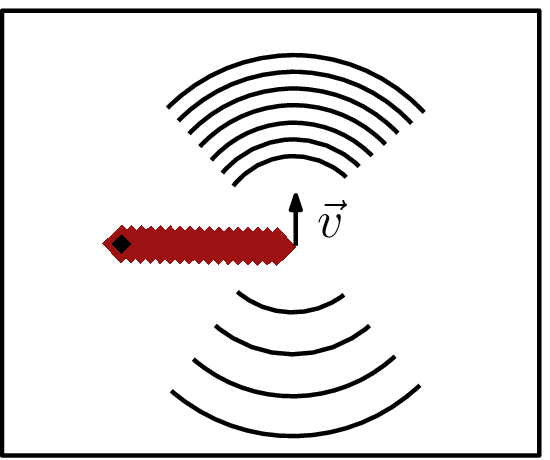}
    \caption{}
    \label{fig:doppler effect}
  \end{subfigure}
  \hfill
  \begin{subfigure}[t]{0.49\linewidth}
    \centering
    \includegraphics[width=3.57cm]{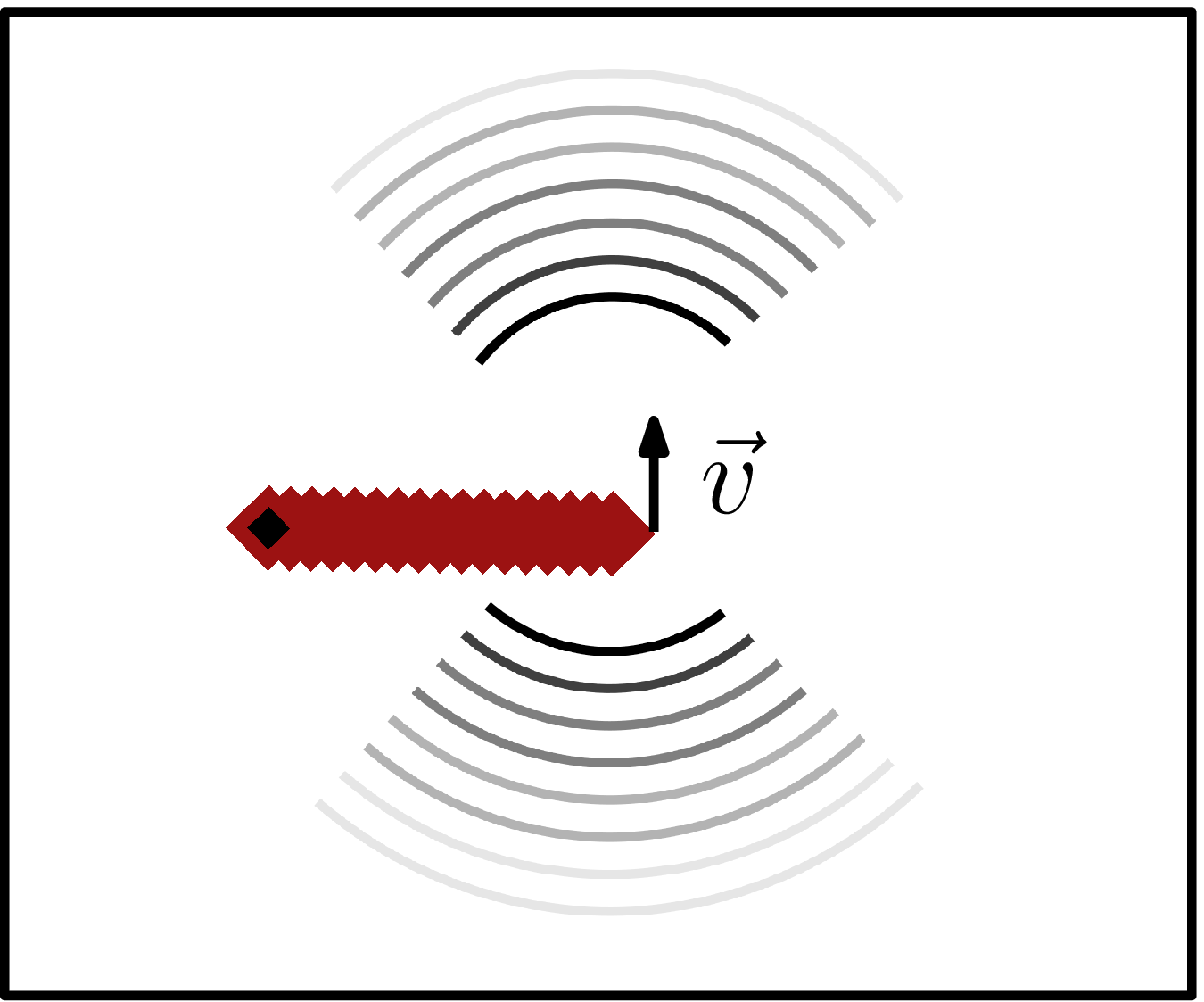}
    \caption{}
    \label{fig:inverse square law}
  \end{subfigure}
  \caption{Different sound properties in the pendulum scenario.
    \ref{fig:doppler effect}) Depicts the Doppler effect. As the sound
    source moves near (away from) the observer, the arrival time of
    the emitted waves decreases (increases), thus increasing
    (decreasing) the frequency. \ref{fig:inverse square law}) Depicts
    the how the amplitude of the sound decreases with the distance
    from the source. Fading semi-circles denote smaller intensities.}
  \label{fig:doppler and amplitude}
\end{figure}

In this scenario, we assume a simple model for the sound
generation. Specifically, we assume that, at each timestep, the
frequency $f'_i$ heard by each sound receiver $\rho_i$ follows the
Doppler effect. The Doppler effect measures the change in frequency
heard by an observer as it moves towards or away from the
source. Slightly abusing our notation, we let $\boldsymbol{\rho}_i$
denote the position of sound receiver $\rho_i$ and $\boldsymbol{e}$
the position of the sound emitter. Formally,
\begin{equation*}
  f'_i
  =
  \left(
    \frac
    {c + \dot{\vect{\rho}}_i \cdot \left( \vect{e} - \vect{\rho}_i \right)}
    {c - \dot{\vect{e}} \cdot \left( \vect{\rho}_i - \vect{e} \right)}
  \right)
  f_0,
\end{equation*}
where $c$ is the speed of sound and we use the dot notation to
represent velocities. Figure~\ref{fig:doppler effect} depicts the
Doppler effect in the pendulum scenario.

We then let the amplitude $a_i$ heard by receiver $\rho_i$ follow the
inverse square law
\begin{equation*}
  a_i = \frac{K}{\|\vect{e} - \vect{\rho}_i\|^2},
\end{equation*}
where $K$ is a scaling constant. Figure~\ref{fig:inverse square law}
depicts the inverse square law applied to the pendulum scenario,
showing how the amplitude of the sound generated decreases with the
distance to the source.

We now provide details on how our approach was set up. All constants
and training hyper-parameters used are summarized in
Appendix~\ref{subsec:constants and params}.

\subsubsection{Learn a perceptual model of the world}
\label{subsubsec:pendulum learn to map inputs}

For this task, we adopted the Associative Variational AutoEncoder
(AVAE) to learn the family of latent mappings $\F$. The AVAE was
trained over a dataset $\D(\I)$ with $M$ observations of images and
sounds
$\vect{i}^m = \left(i_\mathrm{image}^m, i_\mathrm{sound}^m\right)$,
collected using a random controller. The random controller proved to
be enough to cover the state space. Before training, the images were
preprocessed to black and white and resized to $60 \times 60$
pixels. The sounds were normalized to the range $[0, 1]$, assuming the
minimum and maximum values found in the $M$ samples.

For the image-specific encoder we adopted an architecture with two
convolutional layers and two fully connected layers. The two
convolutional layers learned $32$ and $64$ filters, respectively, each
with kernel size $4$, stride $2$ and padding $1$. The two fully
connected layers had $256$ neurons each. Swish
activations~\cite{ramachandran2017arxiv} were used. For the
sound-specific encoder, we adopted an architecture with two fully
connected layers, each with $50$ neurons. One dimension batch
normalization was used between the two layers. The decoders followed
similar architectures. The optimization used Adam gradient with
\textsc{pytorch}'s default parameters, and learning rate
$\eta_{\textsc{avae}}$.

The AVAE loss function penalized poor reconstruction of the image and
sound. Image reconstruction loss was measured by binary cross entropy
scaled by constant $\lambda_{\mathrm{image}}$, and sound
reconstruction loss was measured by mean squared error scaled by
constant $\lambda_{\mathrm{sound}}$. The prior divergence loss terms
were scaled by $\beta$, and the symmetrical KL divergence term by
$\alpha$.

\subsubsection{Learn to act in the world}
\label{subsubsec:pendulum learn to act}

The agent learned how to perform the task using the DDPG algorithm,
while only having access to the image input modality---that is
$\boldsymbol{I}_\mathrm{train} = I_{\mathrm{image}}$. These image
observations are encoded into the latent space using
$\F_\mathrm{train} = \F_\mathrm{image}$---the image-specific encoder
of the AVAE trained in~\ref{subsubsec:pendulum learn to map
  inputs}. Thus, the agent learns a policy $\pi$ that maps latent
states to actions.

The actor and critic networks consisted of two fully connected layers
of $256$ neurons each. The replay buffer was initially filled with
samples obtained using a controller based on the Ornstein-Uhlenbeck
process, with the parameters suggested by~\citet{lillicrap2015arxiv}.
The Adam gradient was used for optimizing both networks, with learning
rates $\eta_{\textsc{critic}}$ and $\eta_{\textsc{actor}}$.

\subsubsection{Transfer policy}

We evaluated the performance of the policy trained
in~\ref{subsubsec:pendulum learn to act}, when the agent only has
access to the sound input modality, \emph{i.e.},
$\boldsymbol{I}_\mathrm{test} = I_{\mathrm{sound}}$.

Given a sound observation, the agent first preprocesses it using the
latent map $\F_\mathrm{test} = \F_\mathrm{sound}$, generating a
multimodal latent state $\vect{z}$---we denote this process as
\textsc{avae\textsubscript{s}}. The agent then uses the policy to
select the optimal action in this latent state.

As a result, we are measuring the zero-shot transfer performance of
policy $\pi$---that is, the ability of the agent to perform its task
while being provided perceptual information that is completely
different from what it saw during the reinforcement learning step,
without any further training. Table~\ref{tab:pendulum crossmodality
  performance} summarizes the transfer performance in terms of average
reward observations throughout an episode of $300$ frames. Our
approach \textsc{avae\textsubscript{s} + ddpg} is compared with two
baselines:
\begin{itemize}
\item \textsc{random} baseline, which depicts the performance of an
  untrained agent. This effectively simulates the performance one
  would expect from a non-transferable policy trained over image
  inputs, and later tested over sound inputs.
\item \textsc{sound ddpg} baseline, a DDPG agent trained directly over
  sound inputs (\emph{i.e.}  the sounds correspond to the
  states). Provides an estimate on the performance an agent trained
  directly over the test input modality may achieve.
\end{itemize}

From Table~\ref{tab:pendulum crossmodality performance}, we conclude
our approach provides the agent with an out-of-box performance
improvement of over $300\%$, when compared to the untrained agent
(non-transferable policy). It is also interesting to observe that the
difference in performance between our agent and \textsc{sound ddpg}
seems small, supporting our empirical observation that the transfer
policy succeeds very often in the task: swinging the pole
up\footnote{We also note that the performance achieved by the
  \textsc{sound ddpg} agent is similar to that reported in the OpenAI
  gym leaderboard for the \textsc{pendulum} scenario with state
  observations as the position and velocity of the pendulum.}.

\subsection{\textsc{hyperhot}}

We consider the \textsc{hyperhot} scenario, a novel top-down shooter
game scenario inspired by the \textsc{space invaders} Atari
game\footnote{We opted to use a custom environment implemented in
  \textsc{pygame}, since the \textsc{space invaders} environment in
  OpenAI gym does not provide access to game state, making it hard to
  generate simulated sounds.}, where the goal of the agent is to shoot
the enemies above, while avoiding their bullets by moving left and
right.

\begin{figure}[t]
  \centering
  \includegraphics[width=2.75cm]{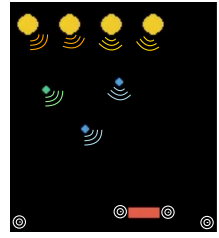}

  \caption{Visual and sound perceptual information in the
    \textsc{hyperhot} scenario. All enemies and bullets emit sounds
    that are received by four microphones at bottom left and right
    ($bl, br$) and paddle left and right ($pl, pr$).}
  \label{fig:hyperhot}
\end{figure}

\begin{table}[b]
  \caption{Zero-shot performance of the policy trained over the image
    input modality, when using sound inputs only. Presents the average
    reward per episode, over $75$ episodes. Results averaged over $10$
    randomly seeded runs.}
  \label{tab:pendulum crossmodality performance}
  \begin{tabular}{@{}m{2.5cm}c@{}}
                                        \multicolumn{2}{c}{\textsc{pendulum}} \\ \toprule
                                        & \ \ \textbf{Rewards}                            \\
  \textbf{Approach}                     & \ \ \textbf{avg} ${\boldsymbol \pm}$ \textbf{std}                       \\ \midrule
  \textsc{avae\textsubscript{s} + ddpg} & $-2.00 \pm 0.97$                    \\ \midrule
  \textsc{random}                       & $-6.30 \pm 0.29$                    \\
  \textsc{sound ddpg}                   & $-1.41 \pm 0.91$                    \\ \bottomrule
\end{tabular}
\end{table}

Similarly to the \textsc{pendulum}, in this scenario, the observations
of the environment include both image and sound components. In
\textsc{hyperhot}, however, the environmental sound is generated by
multiple entities $e_i$ emitting a predefined frequency
$f_0^{(i)}$:
\begin{itemize}
\item
  Left-side enemy units, $e_0$, and right-side enemy units, $e_1$,
  emit sounds with frequencies $f_0^{(0)}$ and $f_0^{(1)}$,
  respectively.
\item
  Enemy bullets. $e_2$, emit sounds with frequency
  $f_0^{(2)}$.
\item
  The agent's bullets, $e_3$, emit sounds with frequency
  $f_0^{(3)}$.
\end{itemize}
The sounds produced by these entities are received by a set of $S$
sound receivers $\left\lbrace \rho_1, \dots, \rho_S
\right\rbrace$. Figure~\ref{fig:hyperhot} depicts the scenario, where
the yellow circles are the enemies; the green and blue bullets are
friendly and enemy fire, respectively; the the agent is in red; and
the sound receivers correspond to the white circles. The agent is
rewarded for shooting the enemies, with the following reward function:

\begin{equation*}
  r =
  \begin{cases}
    10 & \text{if all enemies are killed, \emph{i.e.}, win}          \\
    -1 & \text{if player is killed or time is up, \emph{i.e.}, lose} \\
    0  & \text{otherwise}
  \end{cases}
\end{equation*}
The environment resets whenever the agent collects a non-zero reward,
be it due to winning or losing the game.

We assume the perceptual space of the agent as
$\I = I_{\mathrm{image}} \times I_{\mathrm{sound}}$, with the visual
input modality of the agent, $I_{\mathrm{vision}}$, consisting in the
raw image observation of the environment. The sound, however, is
generated in a more complex and realistic way. We model the sinusoidal
wave of each sound-emitter $e_i$ considering its specific frequency
$f_0^{(i)}$ and amplitude $a_0^{(i)}$. At every frame, we consider the
sound waves of every emitter present in the screen, according to their
distance to each sound receiver in $S$. The sound wave generated by
emitter $e_i$ is observed by receiver $\rho_j$ as

\begin{equation*}
  a^{(i)}
  =
  a_0^{(i)} \exp{\left(-\delta \| \vect{e}_i - \vect{\rho}_j \|^2 \right)}
\end{equation*}
where $\delta$ is a scaling constant, $\vect{e}_i$ and $\vect{\rho}_j$
denote the positions of sound emmitter $e_i$ and sound receiver
$\rho_j$, respectively. We generate each sinusoidal sound wave for a
total of \num{1047} discrete time steps, considering an audio sample
rate of \num{31400} Hz and a video frame-rate of 30 fps. As such, each
sinusoidal sound wave represents the sound heard for the duration of a
single video-frame of the game\footnote{This is similar to what is
  performed in Atari videogames.}. Finally, for each sound receiver,
we sum all emitted waves and encode the amplitude values in 16-bit
audio depth, considering a maximum amplitude value of $a_M$ and a
minimum value of $-a_M$.

We now provide details on how our approach was set up. All constants
and training hyper-parameters used are again summarized in
Appendix~\ref{subsec:constants and params}.

\subsubsection{Learn a perceptual model of the world}

We trained an AVAE model to learn the family of latent mapping $\F$,
with a dataset $\D(\I)$ with $M$ observations of images and sounds
collected using a random controller. Before training, the images were
preprocessed to black and white and resized to $80 \times 80$ pixels,
and the sounds normalized to the range $\left[ 0, 1 \right]$.

For the image-specific encoder we adopted an architecture with three
convolutional layers and two fully connected layers. The three
convolutional layers learned $32, 64$ and $64$ filters,
respectively. The filters were parameterized by kernel sizes $8, 4$
and $2$; strides $4, 2$ and $1$; and paddings $2, 1$ and
$1$. \textsc{ReLU} activations were used throughout. For the
sound-specific component, we used two fully connected layers of $512$
neurons each, with one dimension batch normalization between the
layers. The decoders followed similar architectures.  The increase in
size of these layers when compared to the \textsc{pendulum} task is
due to more complex nature of the sounds considered in this scenario.
The optimizer and loss function were configured in the same way as in
the previous scenario.

\subsubsection{Learn to act in the world}

The agent learned how to play the game using the DQN algorithm, while
having access only to image observations,
$\boldsymbol{I}_\mathrm{train} = I_\mathrm{image}$, corresponding to
the video game frames. The image observations are encoded into the
latent space using $\F_\mathrm{train} = \F_\mathrm{image}$---the
image-specific encoder of the AVAE model trained in the previous
step. As such, the policy learned to play the game, maps these latent
states to actions.

The policy and target networks consisted of two fully connected layers
of $512$ neurons each, and we adopted a decaying $\epsilon$-greedy
policy.

\subsubsection{Transfer policy}

We then evaluated the performance of the policy learned with image
inputs, when the agent only has access to the sound modality,
\emph{i.e.}, $\boldsymbol{I}_\mathrm{test} =
I_{\mathrm{sound}}$. Given a sound observation, the agent preprocesses
it using the latent map $\F_\mathrm{test} = \F_\mathrm{sound}$, thus
generating a multimodal latent state $\vect{z}$---this process is
denoted as \textsc{avae\textsubscript{s}}. The agent then uses the
policy to select the optimal action in this latent state.

Table~\ref{tab:hyperhot crossmodality performance} summarizes the
transfer performance of the policy produced by our approach
\textsc{avae\textsubscript{s} + dqn}, in terms of average discounted
rewards and game win rates over $100$ episodes. We compare the
performance of our approach with additional baselines:

\begin{itemize}
\item \textsc{avae\textsubscript{v} + dqn}, an agent similar to ours,
  but encodes the latent space with visual observations (as
  opposed to sounds).
\item \textsc{image dqn}, a DQN agent trained directly over the visual
  inputs.
\end{itemize}

Considering the results in Table~\ref{tab:hyperhot crossmodality
  performance}, we observe:
\begin{itemize}
\item
  A considerable performance improvement of our approach over the
  untrained agent. The average discounted reward of the
  \textsc{random} baseline is negative, meaning this agent tends to
  get shot often, and rather quickly. This is in contrast with the
  positive rewards achieved by our approach. Moreover, the win rates
  achieved by our approach surpass those of the untrained agent by
  $5$-fold.
\item A performance comparable to that of the agent trained directly
  on the sound, \textsc{sound dqn}. In fact, the average discounted
  rewards achieved by our approach are slightly high\-er. However, we
  note that the \textsc{sound dqn} agent followed the same DQN
  architecture and number of training steps used in our approach. It
  is plausible that with further parameter tuning, the \textsc{sound
    dqn} agent could achieve better performances.
\item The approach that could fine-tune to the most informative
  perceptual modality, \textsc{image dqn}, achieved the highest
  performances. Our approach, while achieving lower rewards, is the
  only able to perform cross-modality policy transfer, that is, being
  able to reuse a policy trained on a different modality. One may
  argue that this trade-off is worthwhile.
\end{itemize}
The DQN networks of all approaches followed similar architectures and
were trained for the same number of iterations.

\begin{table}[b]
  \caption{Zero-shot performance of the policy trained over the image
    modality, when using sound inputs only. Provides a comparison with
    different baselines. Middle column is the average discounted reward
    per episode. Right column is the win rate of the agent. Both
    averaged over $100$ episodes. Results averaged over $10$ randomly
    seeded runs.}
  \label{tab:hyperhot crossmodality performance}
  \begin{tabular}{
  l
  S[separate-uncertainty]
  S[separate-uncertainty]
}
\multicolumn{3}{c}{\textsc{hyperhot}}                                                  \\ \toprule
                        & {\qquad \quad \textbf{Rewards}}       & {\qquad \quad \textbf{Win pct}}       \\
    {\textbf{Approach}} & {\qquad \quad avg $\pm$ std} & {\qquad \quad avg $\pm$ std} \\
    \midrule

       \textsc{avae\textsubscript{s} + dqn} & 0.15 \pm 0.16  & 36.1
                                                               \pm
                                                              10.38
                                                                              \\ \midrule
    \textsc{avae\textsubscript{v} + dqn}    & 0.21 \pm 0.11  & 
                                                               43.20 \pm 7.03 \\
    \textsc{random}                         & -0.33 \pm 0.16 & 8.3
                                                               \pm 5.75       \\
    \textsc{sound dqn}                      & 0.1 \pm 0.22  & 
                                                               27.3\pm 21.44
                                                                              \\
    \textsc{image dqn}                      & 1.54 \pm 0.20  & 75.00 \pm 5.33
                                                                              \\ \bottomrule
\end{tabular}
\end{table}

\subsection{Discussion}
The experimental evaluation performed shows the efficacy and
applicability of our approach. The results show that this approach
effectively enables an agent to learn and exploit policies over
different subsets of input modalities. This sets our work apart from
existing ideas in the literature. For example, DARLA follows a similar
three-stages architecture to allow RL agents to learn policies that
are robust to some shifts in the original
domains~\cite{higgins2017icml}. However, that approach implicitly
assumes that the source and target domains are characterized by
similar inputs, such as raw observations of a camera. This is in
contrast with our work, which allows agents to transfer policies
across different input modalities.

Our approach achieves this by first learning a shared latent
representation that captures the different input modalities. In our
experimental evaluation, for this first step, we used the AVAE model,
which approximates modality-specific latent representations, as
discussed in Section~\ref{subsec:generative model}. This model is
well-suited to the scenarios considered, since these focused on the
transfer of policies trained and reused over distinct input
modalities. We envision other scenarios where training could
potentially take into account multiple input modalities at the same
time. Our approach supports these scenarios as well, when considering
a generative model such as JMVAE~\cite{suzuki2016arxiv}, which can
learn joint modality distributions and encode/decode both modalities
simultaneously.

Furthermore, our approach also supports scenarios where the agent has
access to more than two input modalities. The AVAE model can be
extended to approximate additional modalities, by introducing extra
loss terms that compute the divergence of the new modality specific
latent spaces. However, it may be beneficial to employ generative
models specialized on larger number of modalities, such as the
M$^2$VAE~\cite{korthals2019arxiv}.

\section{Conclusions}
\label{sec:conclusions}

In this paper we explored the use of multimodal latent representations
for capturing multiple input modalities, in order to allow agents to
learn and reuse policies over different modalities. We were
particularly motivated by scenarios of RL agents that learn visual
policies to perform their tasks, and which afterwards, at test time,
may only have access to sound inputs.

To this end, we formalized the multimodal transfer reinforcement
learning problem, and contributed a three stages approach that
effectively allows RL agents to learn robust policies over input
modalities. The first step builds upon recent advances in multimodal
variational autoencoders, to create a generalized latent space that
captures the dependencies between the different input modalities of
the agent, and allow for cross-modality inference. In the second step,
the agent learns how to perform its task over this latent
space. During this training step, the agent may only have access to a
subset of input modalities, with the latent space being encoded
accordingly. Finally, at test time, the agent may execute its task
while having access to a possibly different subset of modalities.

We assessed the applicability and efficacy of our approach in
different domains of increasing complexity. We extended well-known
scenarios in the reinforcement learning literature to include, both
the typical raw image observations, but also the novel sound
components. The results show that the policies learned by our approach
were robust to these different input modalities, effectively enabling
reinforcement learning agents \emph{to play games in the dark}.

\section*{Acknowledgments}

This work was partially supported by national funds through the
Portuguese Funda\c{c}\~{a}o para a Ci\^{e}ncia e a Tecnologia under
project UID/CEC/50021/2019 (INESC-ID multi annual funding) and the
Car\-negie Mellon Portugal Program and its Information and
Communications Technologies Institute, under project
CMUP-ERI/HCI/0051/ 2013. Rui Silva acknowledges the PhD grant
SFRH/BD/113695/2015. Miguel Vasco acknowledges the PhD grant
SFRH/BD/139362/2018.

\newpage

\appendix

\section{Appendix}

\subsection{Constants and hyper-parameters}
\label{subsec:constants and params}

\begin{table}[H]
\begin{tabular}{@{}ccc@{}}
\toprule
                   & $f_0$                  & $440.0$Hz                               \\
                   & $K$                    & $1.0$                                   \\
\sc{scenario}      & $c$                    & $20.0$                                  \\
                   & sound receivers        & $\left\lbrace lb, rb, mt \right\rbrace$ \\
                   & frame stack            & $2$                                     \\ \midrule
                   & latent space           & $10$                                    \\
                   & $\lambda_{\mathrm{image}}, \lambda_{\mathrm{sound}}, \beta,
           \alpha$ & $1.0$                                                            \\
\sc{avae}          & batch size             & 128                                     \\
                   & epochs                 & 500                                     \\
                   & $\eta_{\textsc{avae}}$ & $1\mathrm{e}{-3}$                       \\
                   & $M$                    & \num{20000}                             \\ \midrule
                   & batch size             & 128                                     \\
                   & $\eta_{\textsc{actor}}$, $\eta_{\textsc{critic}}$
                   & $1\mathrm{e}{-4}$, $1\mathrm{e}{-3}$                             \\
\sc{ddpg}          & $\gamma$               & $0.99$                                  \\
                   & max episode length     & 300 frames                              \\
                   & replay buffer          & \num{25000}                             \\
                   & max frames             & \num{150000}
                                                                                      \\
                   & $\tau$                 & $1\mathrm{e}{-3}$                       \\ \bottomrule
\end{tabular}
\caption{Constants used in the \textsc{pendulum} scenario.}
\label{tab:pendulum constants}
\end{table}

\begin{table}[H]
\begin{tabular}{@{}ccc@{}}
\toprule
              & $f_0^{0},\ f_0^{1},\ f_0^{2},\ f_0^{3}$        &
                                                                  $(261,
                                                                  329,
                                                                  392,466)$ Hz          \\
              & $a_0^{0},\ a_0^{1},\ a_0^{2},\ a_0^{3},\ a_M $ & $1.0$                  \\
\sc{scenario} & $\delta$                                       & $0.025$                \\
              & $c$                                            & $20.0$                 \\
              & sound receivers                                &
                                                                  $\left\lbrace
                                                                  lb,
                                                                  rb,
                                                                  pl, pr \right\rbrace$ \\
              & frame stack                                    & $2$                    \\ \midrule
              & latent space                                   & $40$                   \\
              & $\lambda_{\mathrm{image}}$                     & $0.02$
                                                                                        \\
              & $\lambda_{\mathrm{sound}}$                     & $0.015$
                                                                                        \\
              & $\beta$                                        & $1\mathrm{e}{-5}$
                                                                                        \\
\sc{avae}     & $\alpha$                                       & $0.05$                 \\
              & batch size                                     & 128                    \\
              & epochs                                         & 250                    \\
              & $\eta_{\textsc{avae}}$                         & $1\mathrm{e}{-3}$      \\
              & $M$                                            & \num{32000}            \\ \midrule
              & batch size                                     & 128                    \\
              & $\eta$                                         & $1\mathrm{e}{-5}$      \\
\sc{dqn}      & $\gamma$                                       & $0.99$                 \\
              & max episode length                             & 450 frames             \\
              & replay buffer                                  & \num{350000}           \\
              & max frames                                     &
                                                                      \num{1750000}     \\ \bottomrule
\end{tabular}
\caption{Constants used in \textsc{hyperhot} scenario.}
\label{tab:hyperhot constants}
\end{table}

\clearpage

\bibliographystyle{ACM-Reference-Format}
\bibliography{main}

\end{document}